%% file: main.tex
\documentclass[10pt,twocolumn,letterpaper]{article}

\usepackage[pagenumbers]{cvpr} 
\usepackage{multirow}
\usepackage{adjustbox}
\usepackage{nicematrix}
\usepackage{booktabs,makecell}
\usepackage{graphicx,subcaption,calc}
\usepackage{algorithm}
\usepackage{algpseudocode}
\newsavebox{\FigOneABox}
\newsavebox{\FigOneBBox}
\newlength{\FigOneGap}
\newlength{\FigOneAWidth}
\newlength{\FigOneBWidth}
\input{preamble}
\definecolor{cvprblue}{rgb}{0.21,0.49,0.74}
\usepackage[pagebackref,breaklinks,colorlinks,allcolors=cvprblue]{hyperref}

\def\confName{CVPR}
\def\confYear{2027}

\title{PURA: Video Token Pruning via Unified Spatiotemporal Redundancy Assessment}
\author{%
  Yuyao Sun$^{1}$, Hongli Yuan$^{1}$, Shuang Li$^{1}$, Tao Deng$^{1}$, Deqing Wang$^{1}$\\[2pt]
  $^{1}$Beihang University
}
\begin{document}
\maketitle
\input{0_abstract/abstract}

\input{1_introduction/introduction}

\input{2_related_work/related_work}

\input{3_method/method}
\input{4_experiments/experiments}

\input{5_conclusion/conclusion}

{
    \small
    \bibliographystyle{ieeenat_fullname}
    \bibliography{main}
}


\end{document}

%% file: 0_abstract/abstract.tex
\begin{abstract}
Video large language models incur substantial inference costs from long visual token sequences. Existing video token compression methods mitigate these costs by exploiting spatial redundancy within frames and temporal redundancy across frames. However, many approaches impose predefined processing orders or fixed allocation ratios for spatial and temporal compression, limiting their ability to adapt to the varying balance of spatiotemporal redundancy across videos. To address this limitation, we propose PURA, a training-free method that prunes visual tokens through unified spatiotemporal redundancy assessment. For each token, PURA constructs spatial and temporal reference candidates and selects the reference with the lowest representation error across the two dimensions. The representation error is combined with token importance to determine each token's removal cost. To avoid the time overhead of removing the lowest-cost token one by one, we introduce Coordinated Pruning, which iteratively groups mutually compatible low-cost removals into batches for simultaneous execution and updates the pruning state after each batch until the target budget is reached. Extensive experiments on three video large language models and five video understanding benchmarks demonstrate that PURA achieves a favorable trade-off between task performance and inference efficiency.
\end{abstract}

%% file: 1_introduction/introduction.tex
\section{Introduction}
\label{sec:introduction}
\begin{figure}[t]
    \centering
    \vspace{-0.5em}
    \setlength{\FigOneGap}{0.5em}

    \sbox{\FigOneABox}{%
        \includegraphics[height=1cm]{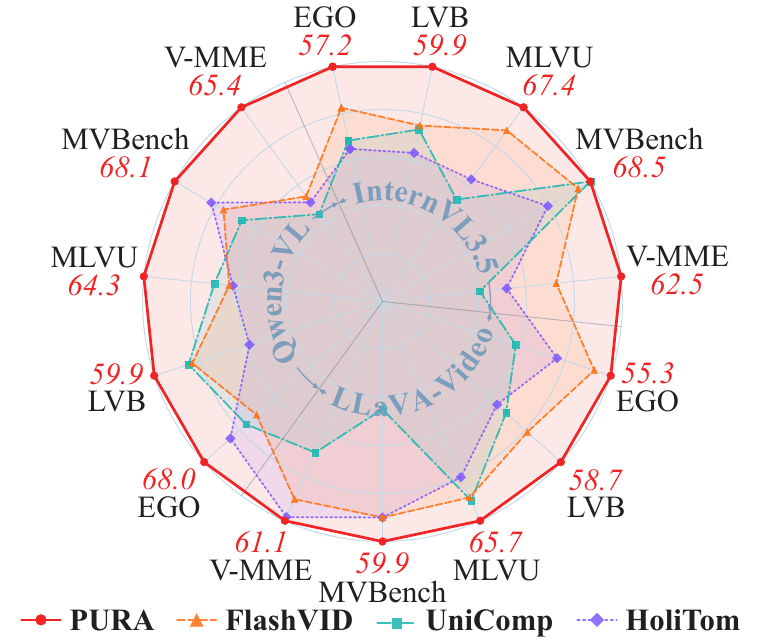}}
    \sbox{\FigOneBBox}{%
        \includegraphics[height=1cm]{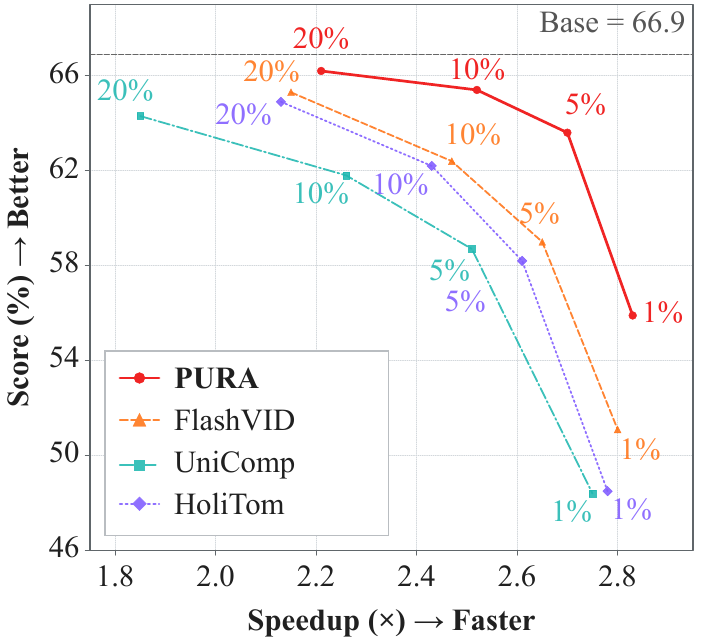}}

    \setlength{\FigOneAWidth}{%
        (\linewidth-\FigOneGap)*
        \ratio{\wd\FigOneABox}{\wd\FigOneABox+\wd\FigOneBBox}}
    \setlength{\FigOneBWidth}{%
        \linewidth-\FigOneGap-\FigOneAWidth}

    \captionsetup[subfigure]{
        font=small,
        skip=0pt,
        justification=centering
    }

    \begin{subfigure}[t]{\FigOneAWidth}
        \centering
        \includegraphics[width=\linewidth]{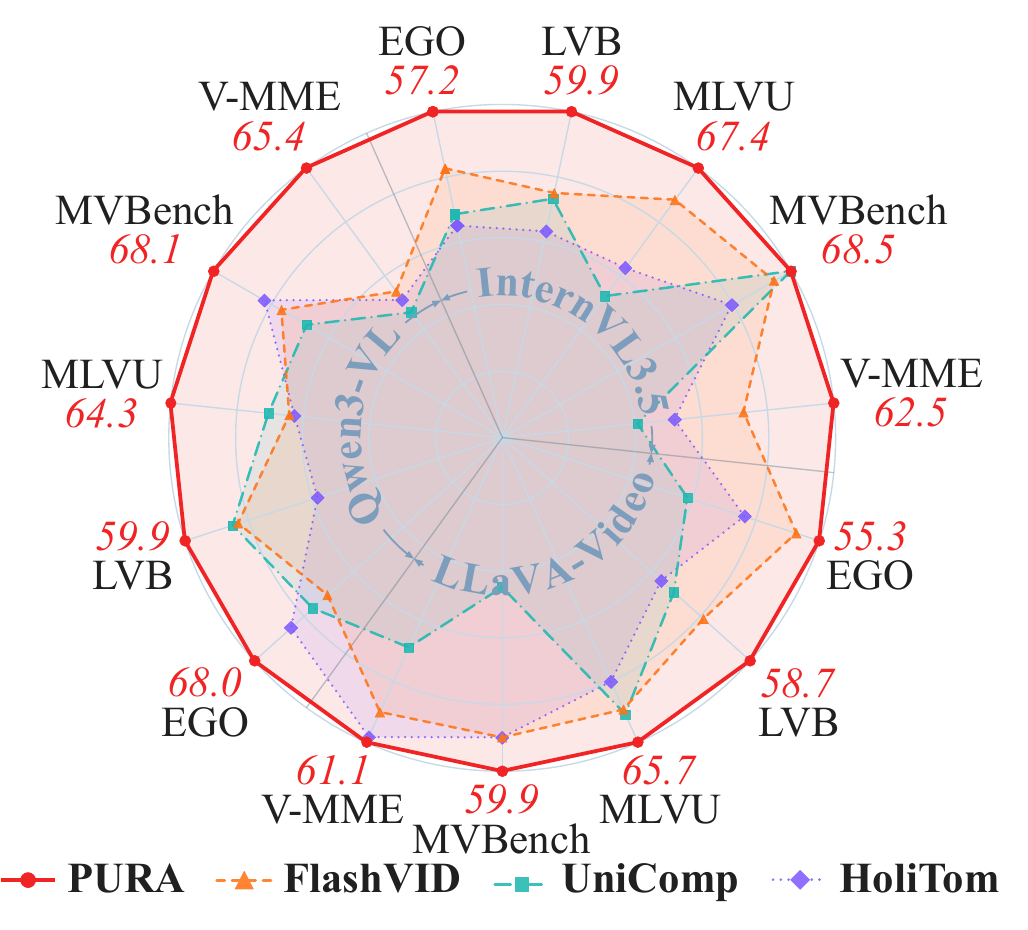}
        \caption{Benchmark performance}
        \vspace{-0.5em}
        \label{fig:intro_performance}
    \end{subfigure}%
    \hspace{\FigOneGap}%
    \begin{subfigure}[t]{\FigOneBWidth}
        \centering
        \includegraphics[width=\linewidth]{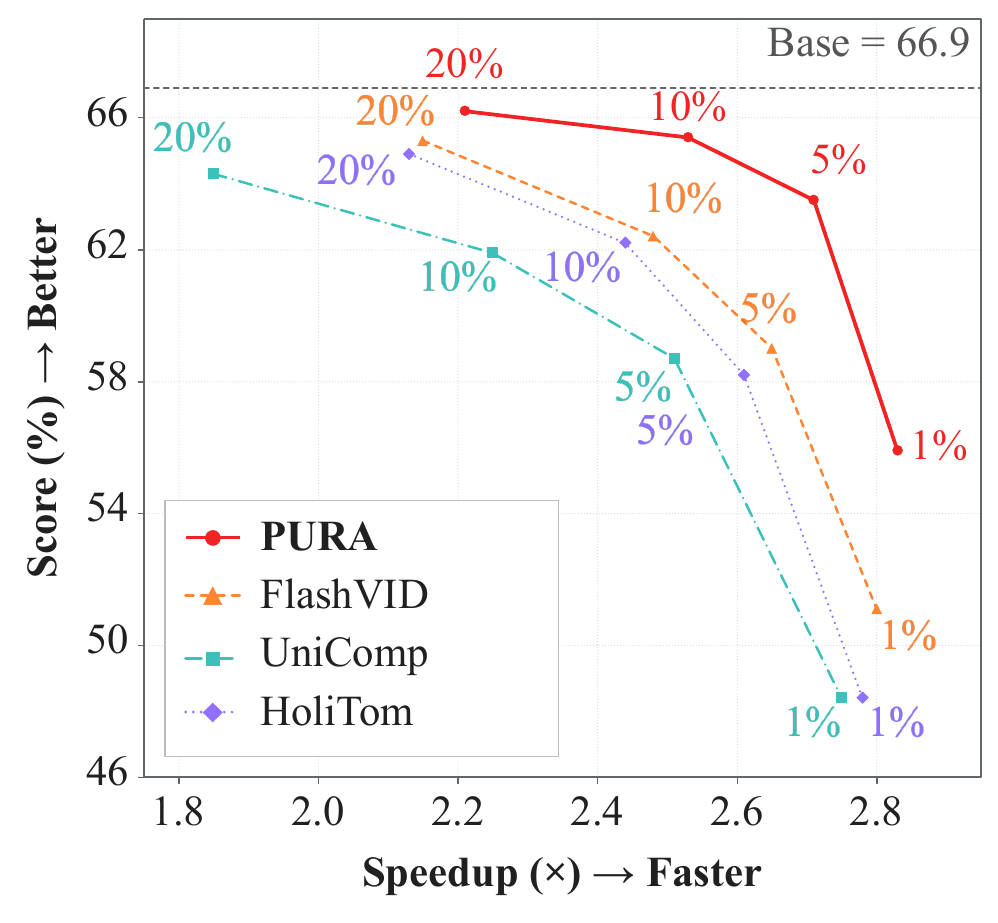}
        \caption{Accuracy--speed trade-off}
        \vspace{-0.5em}
        \label{fig:intro_efficiency}
    \end{subfigure}

    \caption{
        Comparison of performance and inference efficiency across compression methods.
        (a) Performance across benchmarks and models with 10\% visual tokens retained.
        (b) VideoMME accuracy versus end-to-end speedup at different retention
        levels; numbers indicate the retention ratios of visual tokens.
    }
    \label{fig1}
    \vspace{-1em}
\end{figure}


Video large language models (VideoLLMs) have demonstrated strong capabilities in video understanding and reasoning~\cite{videollava,chatunivi,kimivl,videollama3,llavaov2,qwen3vl,internvl3_5}. 
However, processing videos typically yields long visual token sequences from numerous frames, imposing substantial computational and memory costs during LLM inference~\cite{llamavid,dycoke}. These costs grow with the number of input frames and their spatial resolution, making long and visually detailed videos particularly expensive to process.
At the same time, visual token sequences often contain considerable redundant information. Visual token compression has therefore emerged as a promising approach to reducing inference overhead while preserving model performance~\cite{tome,fastv,visionzip,sparsevlm}.


Existing video token compression methods improve efficiency by exploiting spatial redundancy within frames and temporal redundancy across frames. However, many rely on predefined processing orders and manually specified thresholds~\cite{sttm,holitom,prunevid,framefusion}, or preset allocation ratios between frame-wise token selection and cross-frame merging~\cite{fastvid,flashvid}. These predefined strategies limit the flexibility of spatial and temporal compression, making it difficult to adapt their relative contributions to varying redundancy patterns across videos. This can leave redundancy insufficiently reduced while causing unnecessary loss of informative content. 
Recent work~\cite{unified} moves beyond separate spatial and temporal stages through global token selection. Because its matching does not explicitly account for spatiotemporal structure, the method may treat visually similar tokens from different events as redundant.
These limitations motivate a structured yet unified compression strategy to jointly exploit spatiotemporal redundancy in each video.


\begin{figure*}[htbp]
    \vspace{-0.5em}
    \centering
    \includegraphics[width=\linewidth]{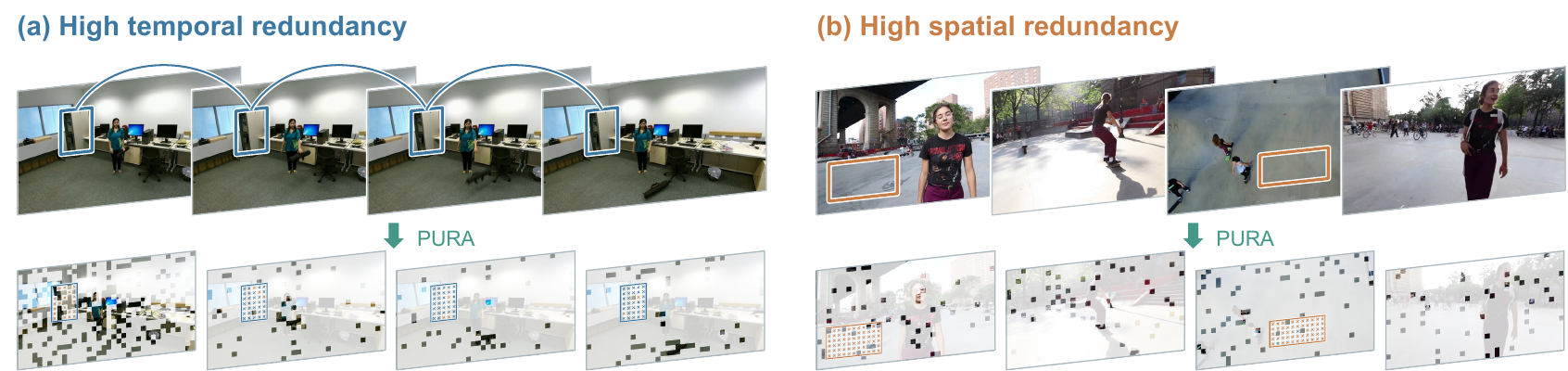}
    \caption{Two representative cases of spatiotemporal redundancy. (a) Repeated visual content across frames exhibits high temporal redundancy. (b) Similar visual content within a frame exhibits high spatial redundancy. PURA jointly models both redundancies, favoring temporal pruning in (a) and spatial pruning in (b). Blue and orange $\times$ markers denote temporal and spatial token removals, respectively.}
    \label{fig2}
    \vspace{-0.5em}
\end{figure*}

To this end, we propose PURA, a training-free method that prunes visual tokens through unified spatiotemporal redundancy assessment. PURA first constructs spatial and temporal reference candidates for each token. Spatial candidates include all other tokens within the same frame, while temporal candidates are obtained by dividing the video into temporal segments, linking each token within each segment to its most similar counterpart in the preceding frame, and tracing these links backward to form a temporal chain. The most similar candidate across the two candidate types is selected as the token's reference, with the feature distance between them measuring the representation error. PURA combines this representation error with token importance, measured by the attention received by the token in the vision encoder, to determine each token's initial removal cost, providing a unified criterion for jointly evaluating both spatial and temporal redundancy.

Given these removal costs, greedily removing the lowest-cost token one at a time and updating the pruning state after each removal is time-consuming, yet directly grouping multiple removals into a batch can be problematic because a token's removal decision relies on its selected reference remaining available. To address this issue, we introduce Coordinated Pruning, which iteratively groups mutually compatible low-cost token removals into batches for simultaneous execution. At each iteration, it preselects a candidate pool of the lowest-cost tokens and scans them in ascending cost order to construct a removal batch, adding a token only if its reference is not in the batch and it is not referenced by any token already selected. These conditions ensure that removing one batch member does not alter the reference or importance of any other member, keeping their removal costs valid for simultaneous removal. After each batch is removed, the importance of removed tokens is transferred to their retained references to preserve the collective semantic contribution of similar tokens. Remaining tokens whose references were removed are then reassigned to their most similar available candidates, and the affected removal costs are recomputed for the next iteration until the target token budget is reached. By executing mutually compatible token removals in batches, Coordinated Pruning substantially reduces ranking and state-update steps while closely approximating sequential greedy pruning.

We evaluate PURA on three video large language models across five video understanding benchmarks. 
Across the tested backbones and retention ratios, PURA consistently achieves the highest average scores. 
Visualizations in Fig.~\ref{fig2} further illustrate how PURA favors temporal pruning when visual content repeats across frames and spatial pruning when similar regions appear within a frame. Meanwhile, Coordinated Pruning cuts mean compression time across benchmarks from 2435.5\,ms to 21.8\,ms, enabling end-to-end speedups of $2.53\times$ at 10\% retention and $2.71\times$ at 5\% retention on VideoMME. Overall, PURA achieves the best trade-off between video understanding performance and inference efficiency among the evaluated methods.

Our contributions are summarized as follows:
\begin{enumerate}
\item We identify a limitation of existing video token compression methods: predefined processing orders and fixed allocation ratios restrict their ability to adapt to varying spatiotemporal redundancy across videos.

\item We propose PURA, a training-free video token compression method through unified spatiotemporal redundancy assessment. It builds structured spatiotemporal references per token and combines representation error with token importance into a unified removal cost for adaptive pruning without predefined orders or fixed ratios.


\item We introduce Coordinated Pruning to accelerate sequential greedy pruning. By grouping mutually compatible low-cost token removals into batches, it substantially reduces sequential ranking and state-update overhead.

\item Extensive experiments across three video large language models and five video understanding benchmarks demonstrate that PURA achieves a superior accuracy--efficiency trade-off over existing methods.
\end{enumerate}

%% file: 2_related_work/related_work.tex
\section{Related Work}
\label{sec:related_work}

\subsection{Video Large Language Models}
Large vision-language models typically connect a vision encoder to a language model through a multimodal projector, enabling joint processing of visual and textual inputs~\cite{instructblip,llava15,minigpt4,vila,blip2}. Most video large language models inherit this architecture, extending visual inputs from individual images to sequences of video frames.
LLaVA-OneVision enables transfer between image and video tasks via unified training on single-image, multi-image, and video data~\cite{llavaonevision}.
LLaVA-Video strengthens video instruction following with synthetic video-language data~\cite{llavavideo}, while InternVL3.5 enhances multimodal reasoning through cascade reinforcement learning~\cite{internvl3_5}.
Alongside these training advances, Qwen2.5-VL improves temporal representation by aligning the temporal component of multimodal rotary position embeddings (MRoPE) with absolute time~\cite{qwen25vl}.
Qwen3-VL further introduces interleaved MRoPE and textual timestamps to improve spatiotemporal modeling and temporal grounding~\cite{qwen3vl}.
As video understanding expands to longer videos and finer visual details, processing more frames at higher resolutions produces longer visual token sequences and increases inference latency and memory consumption, motivating efficient visual token compression.

\subsection{Video Token Compression}
Video token compression exploits redundancy within and across frames to reduce inference costs.
FastVID prunes within temporal segments~\cite{fastvid}, whereas UniComp uses informational uniqueness for frame fusion and token allocation~\cite{unicomp}.
HoliTom and STTM follow predefined merging orders~\cite{holitom,sttm}, while FrameFusion performs cross-frame merging before importance-based pruning~\cite{framefusion}.
FlashVID presets the allocation ratio between frame-wise selection and cross-frame merging~\cite{flashvid}.
However, these methods rely on predefined pipelines or separate budget allocation for different compression operations.
Du et al.~\cite{unified} rank tokens by attention and compare the top $k$ against a global retention pool without explicitly accounting for spatiotemporal structure.
In contrast, PURA constructs structured references from same-frame tokens and temporal correspondences, and defines a unified removal cost using importance-weighted representation error to balance spatial and temporal pruning.
To accelerate this process, Coordinated Pruning executes mutually compatible low-cost removals in batches, transferring importance and updating affected references after each batch to guide later pruning.

%% file: 3_method/method.tex
\section{Method}
\label{sec:method}

\begin{figure*}[t]
    \vspace{-0.5em}
    \centering
    \includegraphics[width=\textwidth]{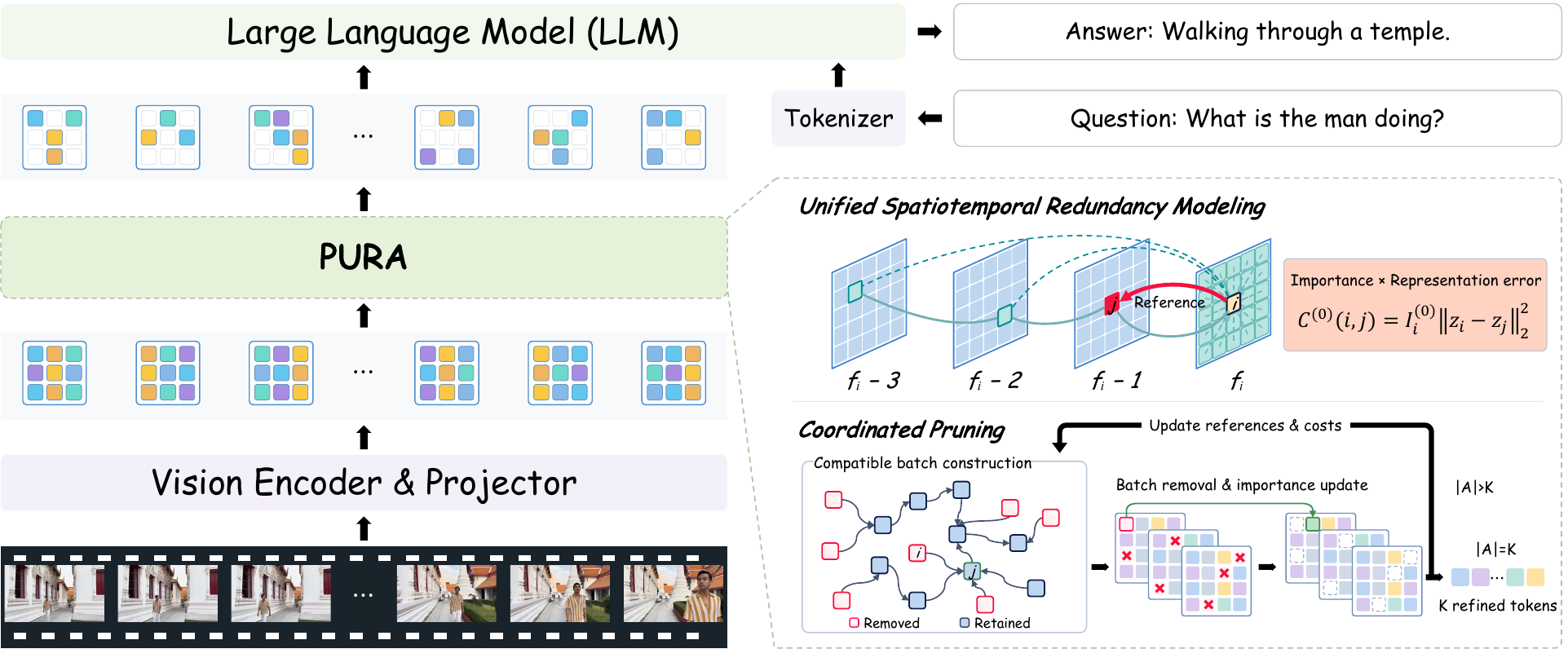}
    \caption{Overview of PURA.}
    \label{fig:method}
    \vspace{-0.5em}
\end{figure*}


\subsection{Overview}
As shown in Figure~\ref{fig:method}, PURA is a training-free method that prunes $N$ visual tokens encoded from a video to a target budget of $K$ tokens for the language model in two stages. \textbf{Unified Spatiotemporal Redundancy Modeling} constructs spatial and temporal candidate sets for each token, selects the most similar candidate as its reference, and weights the representation error by token importance to obtain an initial removal cost. \textbf{Coordinated Pruning} iteratively executes batches of mutually compatible low-cost token removals and updates the pruning state after each batch until $K$ tokens remain. Finally, the retained token representations are refined with information from removed tokens. 

\subsection{Unified Spatiotemporal Redundancy Modeling}
\label{sec}
We first construct the candidate references for each visual token. Let $\mathcal{A}$ denote the set of retained tokens, initialized with all $N$ visual tokens, and let $f_i$ denote the frame index of token $i$. For spatial candidates, we consider all other retained tokens in the same frame as token $i$. To construct temporal candidates, we first obtain a frame embedding by averaging all token embeddings in each frame. We divide the video into temporal segments by placing a boundary wherever the cosine similarity between adjacent frame embeddings falls below a threshold. Within each segment, we construct temporal links between adjacent frames by matching each token in the later frame to the most similar token in a local neighborhood centered on the corresponding spatial position in the preceding frame. Following these temporal links backward from token $i$ forms its temporal chain, and $\mathcal{T}(i)$ denotes the set of earlier tokens reached along this chain. Combining spatial and temporal candidates yields the initial candidate reference set
\begin{equation}
  \mathcal{R}(i) = \{\, j \in \mathcal{A} \mid f_j = f_i,\ j \neq i \,\} \,\cup\, \bigl(\mathcal{T}(i) \cap \mathcal{A}\bigr).
  \label{eq:candidate}
\end{equation}
Temporal links are constructed once and remain fixed throughout pruning, preserving access to earlier retained candidates even when intermediate tokens are removed.

Based on these candidate sets, we then compute the initial removal cost by combining token importance with the representation error relative to a selected reference. Let $z_i$ denote the normalized original feature of token $i$. We select the most similar candidate in $\mathcal{R}(i)$ as its reference $j$ and measure the representation error using the squared feature distance $\|z_i-z_j\|_2^2$. The initial importance $I_i^{(0)}$ is obtained from the attention received by token $i$ in the vision encoder. For encoders with a CLS token, we average CLS-to-token attention across heads. Otherwise, we average the attention weights assigned to token $i$ by other visual tokens across query positions and heads. Weighting the representation error by this importance gives the initial removal cost:
\begin{equation}
    C^{(0)}(i,j)=I_i^{(0)}\|z_i-z_j\|_2^2.
    \label{eq:initial_removal_cost}
\end{equation}
Tokens with an empty $\mathcal{R}(i)$ receive infinite cost and are retained. By selecting the most similar reference, spatial and temporal candidates directly compete for each token, yielding a unified cost that reflects its dominant redundancy.

\subsection{Coordinated Pruning}
\label{sec:pruning}
Given the initial removal costs defined above, a straightforward strategy is to greedily remove the token with the lowest current cost and update the pruning state after each removal. However, reaching the target budget requires $N-K$ sequential steps, making this strategy time-consuming for long token sequences. To reduce this overhead, we introduce Coordinated Pruning, which selects mutually compatible low-cost removals and executes them in batches.


At each iteration, we preselect the $M$ lowest-cost candidates from the remaining tokens based on their current removal costs. We then construct a removal batch $\mathcal{B}$ by scanning these candidates in ascending cost order, starting from an empty batch. A token is added only if (1) its chosen reference is not already in $\mathcal{B}$ and (2) it is not used as a reference by any token already in $\mathcal{B}$. These conditions ensure that removing one batch member does not alter the reference or importance of any other member, keeping their removal costs valid for simultaneous execution. Tokens that fail either condition are skipped and reconsidered in subsequent iterations. The scan stops when all $M$ candidates are examined or $|\mathcal{B}| = \min(B_{\max}, |\mathcal{A}| - K)$ with batch limit $B_{\max}$, after which the selected tokens are removed together. 
Preselecting the $M$ lowest-cost candidates avoids a full sort of the remaining tokens. A larger $B_{\max}$ reduces iterations by permitting more removals before cost updates, which may increase deviation from sequential greedy selection.


After removing the tokens in each batch, we update the importance, references, and removal costs of the remaining tokens to guide the next pruning iteration. Although individual tokens may have low importance, a group of similar tokens can collectively represent semantically important content. We therefore transfer the importance of removed tokens to their selected retained references, so that their importance continues to influence subsequent pruning decisions. Specifically, at iteration $t$, let $\mathcal{B}_j$ denote the tokens in the current batch whose chosen retained reference is $j$. We add their accumulated importance to $j$:
\begin{equation}
I_j^{(t+1)} = I_j^{(t)}
+ \sum\nolimits_{i\in\mathcal{B}_j} I_i^{(t)}.
\label{eq}
\end{equation}
The retained reference thus carries the accumulated importance of the tokens it represents, so subsequent removal costs reflect their combined contribution.


\begin{table*}[ht]
\centering
\setlength{\tabcolsep}{3pt}
\caption{Performance comparison of token compression methods at different retention ratios on Qwen3-VL-8B-Instruct with 64 input frames. Score is the arithmetic mean across EgoSchema (Full), LongVideoBench, MLVU, MVBench, and VideoMME (Avg). Rel. is computed by dividing each benchmark score by its corresponding uncompressed baseline score and averaging the five ratios.}
\label{tab:qwen3_64frames_budgets}

\resizebox{\textwidth}{!}{%
\begin{NiceTabular}{
  @{}
  >{\raggedright\arraybackslash}m{3.65cm}
  >{\centering\arraybackslash}m{1.2cm}
  *{11}{>{\centering\arraybackslash}m{1.15cm}}
  @{}
}
\toprule
\Block[l]{2-1}{Method}
& \Block{2-1}{Retention\\Ratio}
& \multicolumn{2}{c}{EgoSchema}
& \Block{2-1}{LVB}
& \Block{2-1}{MLVU}
& \Block{2-1}{MV\\Bench}
& \multicolumn{4}{c}{VideoMME}
& \multicolumn{2}{c}{Avg} \\
\cmidrule(lr){3-4}
\cmidrule(lr){8-11}
\cmidrule(lr){12-13}
& & Subset & Full & & & &
Short & Medium & Long & Avg & Score & Rel \\
\midrule

Qwen3-VL-8B-Instruct & 100\% & 70.4 & 70.1
& 62.8 & 68.9 & 69.2
& 78.8 & 66.2 & 55.8 & 66.9
& 67.6 & 100.0\% \\
\midrule

HoliTom (NeurIPS'25) & \Block{6-1}{20\%} & 68.0 & 68.4
& 60.6 & 65.3 & 68.4
& 76.7 & 63.1 & 54.9 & 64.9
& 65.5 & 96.9\% \\

FlashVID (ICLR'26) & & 67.4 & 67.8
& 60.6 & 63.9 & 67.3
& 77.2 & 63.0 & 55.7 & 65.3
& 65.0 & 96.2\% \\

Unified (CVPR'26) & & 65.6 & 66.7
& 58.8 & 64.2 & 64.4
& 76.6 & 60.6 & 55.8 & 64.3
& 63.7 & 94.2\% \\

UniComp (CVPR'26) & & 67.2 & 68.1
& 60.4 & 65.1 & 67.9
& 75.6 & 62.7 & 54.6 & 64.3
& 65.2 & 96.4\% \\

V-CAST (NeurIPS'26) & & 67.2 & 67.7
& 60.7 & 65.0 & 67.5
& 76.1 & 62.8 & 54.4 & 64.4
& 65.1 & 96.3\% \\

PURA  & & 68.4 & 69.1
& 62.2 & 66.3 & 68.6
& 78.2 & 64.9 & 55.6 & 66.2
& \textbf{66.5} & \textbf{98.4\%} \\
\midrule

HoliTom (NeurIPS'25) & \Block{6-1}{10\%} & 65.2 & 67.0
& 57.4 & 61.9 & 66.9
& 72.4 & 59.6 & 54.7 & 62.2
& 63.1 & 93.3\% \\

FlashVID (ICLR'26) & & 64.4 & 66.0
& 58.9 & 62.0 & 66.5
& 72.8 & 59.9 & 54.6 & 62.4
& 63.2 & 93.5\% \\

Unified (CVPR'26) & & 62.8 & 65.6
& 58.0 & 62.8 & 62.8
& 73.1 & 58.6 & 53.9 & 61.9
& 62.2 & 92.1\% \\

UniComp (CVPR'26) & & 66.4 & 66.4
& 59.0 & 62.4 & 65.9
& 72.3 & 59.8 & 53.4 & 61.9
& 63.1 & 93.4\% \\

V-CAST (NeurIPS'26) & & 63.4 & 65.4
& 59.7 & 61.9 & 65.2
& 72.8 & 59.7 & 52.0 & 61.5
& 62.7 & 92.9\% \\

PURA  & & 66.6 & 68.0
& 59.9 & 64.3 & 68.1
& 77.3 & 63.6 & 55.2 & 65.4
& \textbf{65.1} & \textbf{96.4\%} \\
\midrule

HoliTom (NeurIPS'25) & \Block{6-1}{5\%} & 59.8 & 63.2
& 55.6 & 57.7 & 64.1
& 68.0 & 55.0 & 51.7 & 58.2
& 59.8 & 88.4\% \\

FlashVID (ICLR'26) & & 61.2 & 63.3
& 55.7 & 58.5 & 63.4
& 68.4 & 55.9 & 52.8 & 59.0
& 60.0 & 88.7\% \\

Unified (CVPR'26) & & 61.8 & 63.8
& 54.0 & 59.0 & 61.4
& 68.8 & 56.1 & 53.3 & 59.4
& 59.5 & 88.0\% \\

UniComp (CVPR'26) & & 60.0 & 62.9
& 55.9 & 58.5 & 63.5
& 67.1 & 56.6 & 52.4 & 58.7
& 59.9 & 88.6\% \\

V-CAST (NeurIPS'26) & & 60.4 & 63.0
& 57.9 & 58.7 & 61.6
& 67.3 & 54.6 & 51.7 & 57.9
& 59.8 & 88.6\% \\

PURA  & & 64.4 & 66.9
& 58.6 & 61.8 & 66.7
& 75.4 & 60.9 & 54.1 & 63.5
& \textbf{63.5} & \textbf{93.9\%} \\
\midrule

HoliTom (NeurIPS'25) & \Block{6-1}{1\%} & 48.0 & 53.5
& 49.3 & 51.6 & 50.4
& 51.1 & 46.7 & 47.6 & 48.4
& 50.6 & 75.0\% \\

FlashVID (ICLR'26) & & 50.0 & 55.2
& 49.6 & 51.8 & 51.6
& 55.0 & 49.2 & 49.2 & 51.1
& 51.9 & 76.8\% \\

Unified (CVPR'26) & & 50.4 & 55.4
& 49.8 & 52.6 & 53.7
& 53.9 & 48.2 & 49.0 & 50.4
& 52.4 & 77.5\% \\

UniComp (CVPR'26) & & 41.6 & 50.6
& 47.3 & 50.8 & 48.8
& 49.9 & 46.8 & 48.4 & 48.4
& 49.2 & 72.8\% \\

V-CAST (NeurIPS'26) & & 50.6 & 56.7
& 52.1 & 53.5 & 52.7
& 56.1 & 48.0 & 47.3 & 50.5
& 53.1 & 78.6\% \\

PURA  & & 56.4 & 61.0
& 53.5 & 55.9 & 60.6
& 64.8 & 52.8 & 50.1 & 55.9
& \textbf{57.4} & \textbf{84.9\%} \\
\bottomrule
\end{NiceTabular}%
}
\end{table*}


Since similarities between tokens remain unchanged during pruning, a chosen reference remains the most similar available candidate as long as it is retained. We therefore select a new reference only when the previous one is removed, choosing the remaining candidate with the highest cosine similarity. We recompute removal costs only for tokens whose importance or chosen reference has changed, using Eq.~\eqref{eq:initial_removal_cost} with the current importance and reference, while reusing the costs of unaffected tokens. Based on the updated removal costs, we then preselect the $M$ lowest-cost remaining candidates for the next pruning iteration. This process repeats until $K$ tokens remain.

After pruning, we further refine each retained token's representation by merging visual information from the corresponding removed tokens to mitigate potential information loss.
Specifically, we follow the recorded removal links from each removed token $i$ to its final retained reference $j$, and include $i$ in the fusion set $\mathcal{F}(j)$ only if the cosine similarity between their features exceeds $0.8$ to ensure semantic consistency.
With $x_i$ denoting the original feature of token $i$, we then compute the refined feature of $j$ as
\begin{equation}
\tilde{x}_j =
\frac{x_j + \sum_{i\in\mathcal{F}(j)} x_i}
{1 + |\mathcal{F}(j)|}.
\label{eq:feature_refinement}
\end{equation}
If $\mathcal{F}(j)$ is empty, $j$ retains its original feature. The resulting $K$ refined visual tokens are fed into the language model.

%% file: 4_experiments/experiments.tex
\section{Experiments}
\label{sec:experiments}

\begin{table*}[t]
\centering
\setlength{\tabcolsep}{3pt}
\caption{Performance comparison of token compression methods on InternVL3.5-8B and LLaVA-Video-7B with 64 input frames.}
\label{tab:internvl_llavavideo_budgets}

\resizebox{\textwidth}{!}{%
\begin{NiceTabular}{
  @{}
  >{\raggedright\arraybackslash}m{3.65cm}
  >{\centering\arraybackslash}m{1.2cm}
  *{11}{>{\centering\arraybackslash}m{1.15cm}}
  @{}
}
\toprule
\Block[l]{2-1}{Method}
& \Block{2-1}{Retention\\Ratio}
& \multicolumn{2}{c}{EgoSchema}
& \Block{2-1}{LVB}
& \Block{2-1}{MLVU}
& \Block{2-1}{MV\\Bench}
& \multicolumn{4}{c}{VideoMME}
& \multicolumn{2}{c}{Avg} \\
\cmidrule(lr){3-4}
\cmidrule(lr){8-11}
\cmidrule(lr){12-13}
& & Subset & Full & & & &
Short & Medium & Long & Avg & Score & Rel \\
\midrule

InternVL3.5-8B & 100\%
& 62.4 & 58.6 & 62.5 & 69.5 & 73.1
& 72.2 & 64.9 & 54.1 & 63.7
& 65.5 & 100.0\% \\
\midrule

HoliTom (NeurIPS'25) & \Block{6-1}{20\%}
& 60.8 & 56.9 & 58.9 & 67.7 & 68.7
& 73.7 & 60.7 & 52.6 & 62.3
& 62.9 & 96.1\% \\

FlashVID (ICLR'26) &
& 60.4 & 57.4 & 59.2 & 67.3 & 66.6
& 74.2 & 62.4 & 53.0 & 63.2
& 62.7 & 96.0\% \\

Unified (CVPR'26) &
& 59.4 & 56.3 & 59.3 & 66.8 & 63.6
& 74.0 & 61.6 & 52.2 & 62.6
& 61.7 & 94.5\% \\

UniComp (CVPR'26) &
& 59.2 & 57.6 & 60.4 & 67.3 & 70.3
& 74.1 & 60.1 & 51.4 & 61.9
& 63.5 & 97.0\% \\

V-CAST (NeurIPS'26) &
& 58.6 & 55.7 & 59.8 & 66.4 & 71.3
& 73.4 & 60.3 & 50.8 & 61.5
& 62.9 & 96.1\% \\

PURA &
& 60.6 & 58.2
& 61.0 & 67.9 & 70.0
& 75.3 & 63.8 & 52.6 & 63.9
& \textbf{64.2} & \textbf{98.1\%} \\
\midrule

HoliTom (NeurIPS'25) & \Block{6-1}{10\%}
& 58.0 & 55.2 & 57.7 & 64.9 & 67.1
& 70.1 & 57.7 & 50.8 & 59.5
& 60.9 & 93.0\% \\

FlashVID (ICLR'26) &
& 58.8 & 56.2 & 58.4 & 66.6 & 68.1
& 72.1 & 58.7 & 51.6 & 60.8
& 62.0 & 94.8\% \\

Unified (CVPR'26) &
& 58.4 & 55.1 & 57.7 & 64.4 & 62.0
& 72.3 & 59.1 & 50.3 & 60.6
& 60.0 & 91.8\% \\

UniComp (CVPR'26) &
& 57.4 & 55.4 & 58.3 & 64.2 & 68.5
& 70.4 & 55.6 & 50.4 & 58.8
& 61.0 & 93.2\% \\

V-CAST (NeurIPS'26) &
& 57.0 & 53.2 & 57.3 & 63.6 & 67.8
& 67.8 & 56.3 & 50.7 & 58.3
& 60.0 & 91.6\% \\

PURA &
& 60.2 & 57.2
& 59.9 & 67.4 & 68.5
& 75.1 & 61.3 & 51.0 & 62.5
& \textbf{63.1} & \textbf{96.5\%} \\
\midrule

LLaVA-Video-7B & 100\%
& 59.6 & 57.2 & 58.9 & 67.3 & 60.4
& 77.3 & 62.4 & 53.4 & 64.4
& 61.6 & 100.0\% \\
\midrule

HoliTom (NeurIPS'25) & \Block{6-1}{20\%}
& 57.4 & 55.9 & 57.2 & 66.3 & 60.8
& 73.9 & 61.2 & 51.2 & 62.1
& 60.5 & 98.1\% \\

FlashVID (ICLR'26) &
& 59.4 & 56.4 & 58.7 & 66.1 & 59.8
& 74.2 & 59.6 & 52.4 & 62.1
& 60.6 & 98.4\% \\

Unified (CVPR'26) &
& 53.0 & 52.7 & 56.6 & 64.6 & 56.9
& 70.3 & 57.7 & 50.2 & 59.4
& 58.0 & 94.1\% \\

UniComp (CVPR'26) &
& 57.6 & 54.8 & 58.3 & 66.5 & 58.1
& 72.2 & 59.7 & 52.2 & 61.4
& 59.8 & 97.0\% \\

V-CAST (NeurIPS'26) &
& 56.8 & 54.9 & 56.0 & 64.3 & 59.5
& 72.7 & 60.4 & 51.2 & 61.4
& 59.2 & 96.1\% \\

PURA &
& 58.6 & 56.6
& 59.8 & 67.1 & 60.6
& 74.6 & 58.3 & 52.7 & 61.9
& \textbf{61.2} & \textbf{99.3\%} \\
\midrule

HoliTom (NeurIPS'25) & \Block{6-1}{10\%}
& 55.2 & 54.0 & 56.6 & 64.4 & 59.3
& 72.6 & 59.0 & 51.4 & 61.0
& 59.1 & 95.8\% \\

FlashVID (ICLR'26) &
& 57.0 & 54.9 & 57.6 & 65.0 & 59.3
& 72.4 & 59.1 & 49.9 & 60.5
& 59.5 & 96.5\% \\

Unified (CVPR'26) &
& 52.0 & 51.3 & 54.9 & 63.6 & 55.9
& 68.6 & 57.3 & 49.2 & 58.4
& 56.8 & 92.1\% \\

UniComp (CVPR'26) &
& 53.4 & 53.0 & 56.9 & 65.1 & 56.6
& 70.3 & 57.3 & 49.9 & 59.2
& 58.2 & 94.3\% \\

V-CAST (NeurIPS'26) &
& 52.4 & 51.7 & 53.0 & 60.3 & 55.7
& 68.0 & 57.6 & 48.4 & 58.0
& 55.7 & 90.4\% \\

PURA &
& 56.6 & 55.3
& 58.7 & 65.7 & 59.9
& 74.3 & 58.2 & 50.8 & 61.1
& \textbf{60.1} & \textbf{97.6\%} \\
\bottomrule
\end{NiceTabular}%
}
\end{table*}

\noindent\textbf{Datasets and Models.}
To evaluate the effectiveness of PURA, we conduct experiments on three representative video LLMs: Qwen3-VL-8B-Instruct~\cite{qwen3vl}, InternVL3.5-8B~\cite{internvl3_5}, and LLaVA-Video-7B~\cite{llavavideo}.
We test all three models across five widely used video understanding benchmarks: VideoMME~\cite{videomme}, LongVideoBench~\cite{longvideobench}, MLVU~\cite{mlvu}, MVBench~\cite{mvbench}, and EgoSchema~\cite{egoschema}. These benchmarks span a wide range of video durations and task types, enabling a comprehensive evaluation of our method.

\noindent\textbf{Compared Methods.}
We compare PURA against five recent video token compression baselines.
HoliTom~\cite{holitom} applies spatiotemporal compression prior to the LLM and merges similar tokens inside the LLM.
FlashVID~\cite{flashvid} performs tree-based spatiotemporal token merging, while UniComp~\cite{unicomp} leverages informational uniqueness to guide token allocation and spatial compression.
V-CAST~\cite{vcast} utilizes trajectory curvature to allocate frame budgets and dual anchors to select spatial tokens.
Finally, we denote the method of Du et al.~\cite{unified} as Unified, which performs global token selection by jointly considering attention scores and similarities to the entire retention pool.

\noindent\textbf{Implementation Details.}
All evaluations use lmms-eval~\cite{lmmseval}, with uniformly sampled video frames.
Following FlashVID~\cite{flashvid}, we adopt its hybrid compression setting and align the layer-averaged token budget for fair comparison.
The initial token importance $I_i^{(0)}$ is computed from the vision encoder's final layer.
Throughout our experiments, we set the temporal segmentation threshold to $0.85$ and use a $7\times7$ neighborhood in the preceding frame for temporal linking.
For Coordinated Pruning, we set $M=1024$ and $B_{\max}=512$.
More details are in the appendix.

\subsection{Main Results}
\label{sec:main_results}
Table~\ref{tab:qwen3_64frames_budgets} compares visual token compression methods on Qwen3-VL-8B-Instruct using 64-frame video inputs across four retention ratios. PURA achieves the highest average score at every retention ratio and consistently ranks first on all five benchmarks. Its advantage becomes more pronounced at lower retention ratios, improving the retained performance over the strongest baseline by 1.5, 2.9, 5.2, and 6.3 percentage points at 20\%, 10\%, 5\%, and 1\% retention, respectively. Moreover, PURA's average score at 5\% retention (63.5) surpasses all baselines at 10\% retention (62.2--63.2), achieving superior performance with only half the retained visual tokens. 
Notably, PURA substantially outperforms Unified across all four retention ratios, supporting the use of structured spatiotemporal references over unconstrained global token matching.

Table~\ref{tab:internvl_llavavideo_budgets} further evaluates PURA on InternVL3.5-8B and LLaVA-Video-7B using 64-frame video inputs at 20\% and 10\% visual token retention.
PURA continues to achieve the highest average score across both models and retention ratios.
On InternVL3.5-8B, PURA retains 98.1\% and 96.5\% of the uncompressed performance at 20\% and 10\% retention, outperforming the strongest baselines by 1.1 and 1.7 percentage points in retained performance, respectively.
On LLaVA-Video-7B, PURA retains a near-lossless 99.3\% of the uncompressed performance at 20\% retention and 97.6\% at 10\%, exceeding the strongest baselines by 0.9 and 1.1 percentage points, respectively.
Overall, PURA delivers consistent gains, demonstrating strong generalization across diverse model families and retention ratios.

\begin{table*}[t]
\centering
\caption{Performance comparison with token compression methods under higher frame counts on Qwen3-VL-8B-Instruct.}
\label{tab:qwen3_more_frames}
\resizebox{\textwidth}{!}{%
\begin{NiceTabular}{l*{12}{c}}
\toprule
\Block[l]{2-1}{Method}
& \Block{2-1}{Frames}
& \Block{2-1}{Retention\\Ratio}
& \multicolumn{2}{c}{EgoSchema}
& \Block{2-1}{LVB}
& \Block{2-1}{MLVU}
& \multicolumn{4}{c}{VideoMME}
& \multicolumn{2}{c}{Avg} \\
\cmidrule(lr){4-5}
\cmidrule(lr){8-11}
\cmidrule(lr){12-13}
& & & Subset & Full & & &
Short & Medium & Long & Avg & Score & Rel \\
\midrule
Base & 64 & 100\%
& 70.4 & 70.1 & 62.8 & 68.9
& 78.8 & 66.2 & 55.8 & 66.9
& 67.2 & 100.0\% \\
\midrule
FlashVID & 256 & 25\%
& 68.4 & 69.1 & 63.4 & 72.8
& 80.6 & 68.1 & 59.4 & 69.4
& 68.7 & 102.2\% \\
V-CAST & 256 & 25\%
& 70.0 & 69.9 & 64.4 & 72.9
& 80.2 & 68.8 & 59.2 & 69.4
& 69.2 & 103.0\% \\
PURA & 256 & 25\%
& 71.4 & 70.8 & 65.8 & 75.4
& 80.4 & 69.1 & 61.4 & 70.3
& \textbf{70.6} & \textbf{105.1\%} \\
\midrule
FlashVID & 512 & 12.5\%
& 68.8 & 69.0 & 63.5 & 73.0
& 78.4 & 67.8 & 60.1 & 68.8
& 68.6 & 102.1\% \\
V-CAST & 512 & 12.5\%
& 68.0 & 68.8 & 65.0 & 73.3
& 77.1 & 68.3 & 59.1 & 68.2
& 68.8 & 102.5\% \\
PURA & 512 & 12.5\%
& 71.6 & 71.1 & 66.3 & 75.5
& 80.0 & 69.9 & 63.4 & 71.1
& \textbf{71.0} & \textbf{105.7\%} \\
\bottomrule
\end{NiceTabular}
}
\end{table*}

\begin{table*}[t]
\centering
\setlength{\tabcolsep}{3pt}
\begin{minipage}[t]{0.48\textwidth}
\centering
\caption{Component ablation on Qwen3-VL-8B-Instruct.}
\label{tab:progressive_qwen}
\resizebox{\linewidth}{!}{%
\begin{NiceTabular}{@{}l*{5}{c}@{}}
\toprule
Variant & EgoSchema & LVB & MVBench & VideoMME & Avg \\
\midrule
Spatial baseline
& 63.6 & 56.6 & 61.7 & 60.1 & 60.5 \\
+ Attention
& 65.2 & 56.7 & 64.1 & 61.3 & 61.8 \\
+ Importance transfer 
& 66.3 & 58.0 & 64.9 & 61.8 & 62.8 \\
+ Merge
& 66.5 & 58.1 & 65.3 & 62.0 & 63.0 \\
+ Temporal (Full)
& 66.9 & 58.6 & 66.7 & 63.5 & 63.9 \\
\bottomrule
\end{NiceTabular}%
}
\end{minipage}
\hfill
\begin{minipage}[t]{0.48\textwidth}
\centering
\caption{Component ablation on LLaVA-Video-7B.}
\label{tab:progressive_llavavideo}
\resizebox{\linewidth}{!}{%
\begin{NiceTabular}{@{}l*{5}{c}@{}}
\toprule
Variant & EgoSchema & LVB & MVBench & VideoMME & Avg \\
\midrule
Spatial baseline      & 52.5 & 55.6 & 57.6 & 58.7 & 56.1 \\
+ Attention           & 55.2 & 55.1 & 59.0 & 60.2 & 57.4 \\
+ Importance transfer & 54.9 & 57.1 & 59.5 & 59.7 & 57.8 \\
+ Merge               & 54.5 & 57.7 & 59.9 & 60.4 & 58.1 \\
+ Temporal (Full)     & 55.3 & 58.7 & 59.9 & 61.1 & 58.8 \\
\bottomrule
\end{NiceTabular}%
}
\end{minipage}
\end{table*}


\subsection{Frame Scaling under a Fixed Token Budget}
\label{sec:more_frames} 

Table~\ref{tab:qwen3_more_frames} evaluates how PURA scales to more input frames on Qwen3-VL-8B-Instruct.
We test 256 and 512 input frames at 25\% and 12.5\% retention, respectively, 
keeping the visual token budget processed by the LLM consistent with the uncompressed 64-frame baseline.
At 256 frames, all three methods outperform the 64-frame baseline on average, with PURA leading at 105.1\% of the baseline performance.
When the input is expanded to 512 frames, FlashVID and V-CAST show no further overall gains, with their relative performance dipping slightly from 102.2\% to 102.1\% and from 103.0\% to 102.5\%, respectively.
In contrast, PURA continues to improve, reaching 105.7\% of the baseline performance.
This advantage is particularly pronounced in long-video understanding, where denser temporal sampling helps capture informative events that sparse sampling may miss.
By effectively utilizing this temporal information, PURA with 512 frames achieves 113.6\% of the uncompressed 64-frame baseline's score on VideoMME Long, exceeding the 107.7\% and 105.9\% achieved by FlashVID and V-CAST, respectively.
These results suggest that jointly modeling spatial and temporal redundancy helps PURA adapt to the increased cross-frame redundancy in denser frame inputs.


\subsection{Component Ablation}
\label{sec:component_ablation}
Tables~\ref{tab:progressive_qwen} and~\ref{tab:progressive_llavavideo} present progressive component ablations across backbones and budgets on Qwen3-VL-8B (5\% retention) and LLaVA-Video-7B (10\% retention).
Starting from a spatial-only baseline based solely on representation error, we successively add attention-based importance weighting, importance transfer, feature merging, and temporal pruning, with each variant retaining the preceding components.
Attention-based importance weighting improves the average score by 1.3 points on both backbones, confirming that attention weights effectively reflect the relative importance of visual tokens.
Building on this configuration, importance transfer brings further gains of 1.0 and 0.4 points on Qwen3-VL-8B and LLaVA-Video-7B, respectively, showing that transferring importance to retained references preserves the collective contribution of similar tokens.
Feature merging subsequently adds 0.2 and 0.3 points, respectively, by fusing removed tokens into their retained references to mitigate information loss.
Finally, incorporating temporal pruning boosts the average score by 0.9 and 0.7 points, respectively, highlighting the benefit of unified spatiotemporal redundancy modeling in video token compression.

\subsection{Inference Efficiency Analysis}

\begin{table*}[t]
\centering
\caption{Efficiency on VideoMME using Qwen3-VL-8B-Instruct with 64 input frames. All times are reported in ms.}
\label{tab:efficiency}
\setlength{\tabcolsep}{8pt}
\resizebox{\textwidth}{!}{%
\begin{NiceTabular}{@{}l*{8}{c}@{}}
\toprule
Method & Retention Ratio & TFLOPs & Compression & LLM Prefill & LLM Total & End-to-End Latency & Score & Rel \\
\midrule
Vanilla & 100.0\% & 548.9 & 0 & 2385
& 2446 & 3521 & 66.9 & 100.0\% \\

HoliTom & 10.0\% & 134.9 & 74 & 231 ($10.32\times$)
& 288 & 1445 ($2.44\times$) & 62.2 & 93.0\% \\

FlashVID & 10.0\% & 143.2 & 52 & 223 ($10.70\times$)
& 282 & 1417 ($2.48\times$) & 62.4 & 93.3\% \\

Unified & 10.0\% & 142.7 & 509 & 213 ($11.20\times$)
& 277 & 1870 ($1.88\times$) & 61.9 & 92.5\% \\

UniComp & 10.0\% & 142.4 & 216 & 210 ($11.36\times$)
& 272 & 1563 ($2.25\times$) & 61.9 & 92.5\% \\

V-CAST & 10.0\% & 142.7 & 8 & 214 ($11.14\times$)
& 280 & 1362 ($2.59\times$) & 61.5 & 91.9\% \\

PURA & 10.0\% & 142.8 & 31 & 221 ($10.79\times$)
& 278 & 1393 ($2.53\times$) & \textbf{65.4} & \textbf{97.8\%} \\

PURA & 5.0\% & 126.7 & 29 & 131 ($18.21\times$)
& 187 & 1299 ($2.71\times$) & \textbf{63.5} & \textbf{94.9\%} \\
\bottomrule
\end{NiceTabular}%
}
\vspace{-1em}
\end{table*}

Table~\ref{tab:efficiency} evaluates inference efficiency on VideoMME using Qwen3-VL-8B-Instruct with 64 frames. All measurements are taken on a single NVIDIA RTX 5090 GPU with a batch size of 1.
The results show that PURA achieves a favorable accuracy--latency trade-off with modest compression overhead.
At a 10\% token retention ratio, PURA reduces prefill latency to 221 ms and end-to-end latency to 1393 ms for $10.79\times$ and $2.53\times$ speedups over Vanilla, retaining 97.8\% of Vanilla's score at 65.4 and achieving both lower overall latency and higher accuracy than most competing baselines.
This latency advantage largely stems from its 31 ms compression stage, which is faster than FlashVID (52 ms), HoliTom (74 ms), and UniComp (216 ms), and $16.4\times$ faster than global pool selection in Unified (509 ms).
Relative to V-CAST, the fastest competing baseline, PURA improves the VideoMME score by 3.9 points with only 2.3\% additional end-to-end latency.
At a 5\% token retention ratio, PURA further reduces prefill latency to 131 ms and overall latency to 1299 ms, achieving $18.21\times$ and $2.71\times$ speedups while maintaining a VideoMME score of 63.5.
Despite retaining half as many tokens, PURA achieves higher accuracy and lower end-to-end latency than all competing compression methods evaluated at 10\% retention.

\begin{table}[t]
\centering
\caption{
Ablation of reference candidates, spatiotemporal budget allocation, and pruning order on Qwen3-VL-8B-Instruct.
}
\label{tab:spatiotemporal_ablation}
\setlength{\tabcolsep}{3pt}
\resizebox{\columnwidth}{!}{%
\begin{tabular}{@{}lccccc@{}}
\toprule
Variant & EgoSchema & LVB & MVBench & VideoMME & Avg \\
\midrule

Spatial-Only
& 66.5 & 58.1 & 65.3 & 62.0 & 63.0 \\
\midrule
\multicolumn{6}{@{}l@{}}{
    \textit{Reference candidates}
} \\
Global
& 65.7 & 58.3 & 62.8 & 62.2 & 62.3 \\
Segment-Wide
& 66.1 & 58.9 & 65.2 & 62.7 & 63.2 \\

\midrule
\multicolumn{6}{@{}l@{}}{
    \textit{Fixed budget allocation}
} \\
Global-Level
& 66.8 & 57.7 & 64.9 & 61.8 & 62.8 \\
Benchmark-Level
& 66.7 & 57.9 & 64.7 & 62.3 & 62.9 \\

\midrule
\multicolumn{6}{@{}l@{}}{
    \textit{Staged pruning}
} \\
Temporal-First
& 66.7 & 58.3 & 65.6 & 62.1 & 63.2 \\
Spatial-First
& 66.1 & 58.2 & 63.4 & 62.5 & 62.6  \\

\midrule
\textbf{PURA}
& 66.9 & 58.6 & 66.7 & 63.5 & 63.9 \\

\bottomrule
\end{tabular}%
}
\end{table}

\subsection{Ablation of Reference Candidates, Spatiotemporal Allocation, and Pruning Order}
Table~\ref{tab:spatiotemporal_ablation} evaluates reference candidates, spatiotemporal allocation, and pruning order on Qwen3-VL-8B-Instruct at 5\% token retention, with all variants sharing the same removal cost and state-update rules.
We also include Spatial-Only (63.0), which disables temporal references in full PURA.
For reference candidates, Global considers all other retained tokens across the video, akin to the global candidate pool in Unified~\cite{unified}, whereas Segment-Wide augments the same-frame candidates with all retained tokens from every other frame in the same segment.
Notably, Global (62.3) falls 0.7 points below Spatial-Only, whereas PURA achieves 63.9, outperforming Global by 1.6 points and confirming the benefit of structured spatiotemporal references over unconstrained global candidates.
PURA also outperforms Segment-Wide (63.2) by 0.7 points while requiring fewer temporal similarity computations through sparse local chains, achieving better pruning performance with lower matching complexity.
To examine spatiotemporal allocation, we record each video's temporal removal share $\rho_v$ in full PURA, defined as the fraction of token removals that use temporal references.
Global Level and Benchmark Level average $\rho_v$ across all videos and within each benchmark, respectively, to set fixed removal quotas. Both variants use joint cost-based pruning until either quota is met, then restrict further removals to the other dimension.
Their mean scores of 62.8 and 62.9 fall slightly below Spatial-Only and trail PURA by 1.1 and 1.0 points, respectively, highlighting the benefit of adapting removal proportions to each video.
To examine pruning order, Temporal First and Spatial First use PURA's exact per-video spatial and temporal removal quotas, but exhaust one quota before switching to the other dimension.
Even with matched quotas, Spatial First (62.6) falls below Spatial-Only, and PURA outperforms Temporal First (63.2) and Spatial First by 0.7 and 1.3 points, respectively, highlighting the benefit of jointly pruning tokens with low removal cost across both dimensions.


\begin{table}[t]
\centering
\caption{Ablation of batch limit $B_{\max}$ in Coordinated Pruning on Qwen3-VL-8B-Instruct. Results averaged over five benchmarks.}
\label{tab:batch_fivebench_avg}
\small
\setlength{\tabcolsep}{3pt}
\begin{tabular*}{\linewidth}
{@{\extracolsep{\fill}}lrrrr@{}}
\toprule
Batch & Time & Speedup
& Token diff. & Score \\
\midrule
Greedy & 2435.5 & 1.0 & 0.0\% & 63.3 \\
\midrule
32 & 97.8 & 24.9 & 0.0\% & 63.3 \\
64 & 58.2 & 41.8 & 0.0\% & 63.4 \\
128 & 37.9 & 64.3 & 0.1\% & 63.3 \\
256 & 27.1 & 89.9 & 0.3\% & 63.3 \\
512 & 21.8 & 111.7 & 1.1\% & 63.5 \\
1024 & 19.0 & 128.2 & 3.3\% & 63.4 \\
\bottomrule
\end{tabular*}
\end{table}

\subsection{Ablation of Pruning Batch Limit}
Table~\ref{tab:batch_fivebench_avg} reports the effect of the batch limit $B_{\max}$ in Coordinated Pruning on Qwen3-VL-8B-Instruct at a 5\% token retention ratio, averaged over all five benchmarks.
Larger batch limits accelerate token compression but increase selection deviation from sequential greedy pruning, as measured by the proportion of non-overlapping retained tokens.
Nevertheless, the mean accuracy remains highly stable from 63.3\% to 63.5\% across all tested batch limits.
At $B_{\max}=512$, compression latency drops from 2435.5\,ms to 21.8\,ms, achieving a 111.7$\times$ speedup with only a 1.1\% token difference and a slight accuracy gain to 63.5\%.
Further increasing $B_{\max}$ to 1024 marginally reduces latency to 19.0\,ms while tripling the token difference to 3.3\%.
Given these diminishing returns, we set $B_{\max}=512$ by default to balance compression efficiency and selection consistency.

%% file: 5_conclusion/conclusion.tex
\section{Conclusion}
\label{sec:conclusion}
In this work, we propose PURA, a training-free framework for video token compression via unified spatiotemporal redundancy assessment. By jointly evaluating structured spatial and temporal references with importance-weighted representation error, PURA adaptively balances spatial and temporal pruning without predefined orders or fixed quotas. Moreover, Coordinated Pruning removes mutually compatible tokens in batches, greatly reducing sequential ranking and update overhead. Experiments across three VideoLLMs and five benchmarks show that PURA consistently outperforms prior methods, especially under low token budgets. Overall, PURA achieves a favorable trade-off between task performance and inference efficiency.